\documentclass[11pt]{article}

\usepackage{afterpage}
\usepackage[final]{acl}

\usepackage{latexsym}

\iftutex
\usepackage{fontspec}
\ifdefined\directlua
\newfontfamily\tamilfont{FreeSerif.otf}[Renderer=HarfBuzz,Script=Tamil,AutoFakeBold=2.0,Scale=0.92]
\else
\newfontfamily\tamilfont{FreeSerif.otf}[Script=Tamil,AutoFakeBold=2.0,Scale=0.92]
\fi
\ifdefined\directlua
\newfontfamily\nepalifont{FreeSerif.otf}[Renderer=HarfBuzz,Script=Devanagari,Scale=0.92]
\else
\newfontfamily\nepalifont{FreeSerif.otf}[Script=Devanagari,Scale=0.92]
\fi
\newcommand{\netext}[1]{{\nepalifont #1}}
\newcommand{\tatext}[1]{{\tamilfont #1}}
\else
\newcommand{\netext}[1]{[Nepali text: compile with XeLaTeX/LuaLaTeX]}
\newcommand{\tatext}[1]{[Tamil text: compile with LuaLaTeX]}
\usepackage{times}

\usepackage[T1]{fontenc}

\usepackage[utf8]{inputenc}
\fi
\DeclareRobustCommand{\ub}[1]{\leavevmode{\setbox0\hbox{#1}\ooalign{\copy0\cr\hidewidth\raisebox{-.35ex}{\rule{.75\wd0}{.06ex}}\hidewidth\cr}}}

\usepackage{microtype}

\iftutex
\else
\usepackage{inconsolata}
\fi

\usepackage{graphicx}
\usepackage{tikz}

\author{
 \textbf{Deevyanshu Malu\textsuperscript{1*}} \quad
 \textbf{Deeptanshu Malu\textsuperscript{1*}} \quad
 \textbf{Sunita Sarawagi\textsuperscript{1}}
\\
\textsuperscript{*}Equal Contribution \quad \textsuperscript{1}IIT Bombay
\\
 \texttt{\{deevyanshumalu, deeptanshumalu\}@cse.iitb.ac.in}
}

\usepackage{amsmath, amssymb, amsthm}
\usepackage{algorithm}
\usepackage{cleveref}
\usepackage{mathtools}
\usepackage{booktabs}
\usepackage{defs}
\RequirePackage[inline,shortlabels]{enumitem}

\usepackage{latexsym}
\usepackage{amsmath}
\usepackage{dsfont}
\usepackage{amssymb}
\usepackage{mathtools}
\usepackage{subcaption}
\usepackage{fvextra}
\DefineVerbatimEnvironment{wrapverb}{Verbatim}{breaklines=true}
\usepackage{multirow}
\usepackage{defs}
\RequirePackage[inline,shortlabels]{enumitem}
\setlist{nosep}
\usepackage{graphicx}
\usepackage{xcolor}
\usepackage{svg}

\usepackage[most]{tcolorbox}

\newtcolorbox{promptbox}{
    colback=gray!8,
    colframe=gray!50,
    boxrule=0.3pt,
    arc=2pt,
    left=6pt,
    right=6pt,
    top=4pt,
    bottom=4pt
}

\usepackage{xcolor}
\usepackage{colortbl}
\newcommand{\uterm}[1]{\underline{\textbf{#1}}}
\newcommand{\xterm}[1]{\textcolor{red!70!black}{\textbf{#1}}}
\newcommand{\cmark}{\textcolor{green!50!black}{\ding{51}}}
\newcommand{\xmark}{\textcolor{red!70!black}{\ding{55}}}
\usepackage{pifont}  

\usepackage[disable]{todonotes}
\newcommand{\sysname}{RareCipherFT}
\RequirePackage[inline,shortlabels]{enumitem}
\setlist{nosep}
\usetikzlibrary{decorations.pathreplacing,positioning,arrows.meta,fit,backgrounds,shapes.geometric}
\title{Training with Ciphers to Enhance In-Context Terminology Learning For Translation from Rare Languages}

\begin{document}
\maketitle
\newcommand{\sstodo}[1]{\todo[color=cyan]{#1}}
\newcommand{\Ddom}{D_\tau}
\newcommand{\Dc}{D_\text{cipher}}
\newcommand{\Vrare}{\mathcal{V}_{\mathrm{rare}}}
\newcommand{\Vcip}{\mathcal{V}_{\mathrm{cip}}}
\newcommand{\pmin}{p_{\min}}
\newcommand{\Lcipher}{\mathcal{L}_{\mathrm{cipher}}}
\newcommand{\yp}[1]{\vy_{1:#1}}
\newcommand{\yt}{y^*}
\newcommand{\alignT}[2]{{#1}_{A(#2)}}
\newcommand{\model}{P_\theta}
\newcommand{\sep}{\vs}
\newcommand{\prompt}{I}
\newcommand{\sos}{\text{S}}
\newcommand{\fsft}{IC-Train}
\newcommand{\ictrain}{\fsft}
\newcommand{\vxt}{\vx^*}
\newcommand{\vyt}{\vy^*}
\newcommand{\cvyt}{\Tilde{\vy}^*}
\newcommand{\base}{Base}
\newcommand{\fullzeroshot}{ZSFT}
\newcommand{\fullfewshot}{FSFT-Few-Random}
\newcommand{\fullfewshotse}{FSFT-Similar}
\newcommand{\fullfewshotpara}{FSFT-Paraphrase}
\newcommand{\zeroshot}{ZeroShotFT}

\newcommand{\fsrand}{Random-Context}
\newcommand{\fssim}{Similar-Context}
\newcommand{\zs}{Zero-Context}
\newcommand{\fsour}{\ours}

\newcommand{\fsonenear}{One-Near-Context}

\newcommand{\fsnse}{\fsrand}
\newcommand{\fsse}{\fssim}
\newcommand{\fspara}{Para}
\newcommand{\shuffle}{FS-Shuffle}
\newcommand{\ours}{Contrastive-Context}
\renewcommand{\cM}{P_\theta}
\renewcommand{\choose}{\texttt{Choose}}

\newcommand{\simS}{\text{sim}}
\newcommand{\vyh}{\hat{\vy}}
\newcommand{\copyscore}{\text{Copy-score}}
\newcommand{\iclscore}{\text{ICL-score}}
\newcommand{\iwlscore}{\text{IWL-score}}
\newcommand{\method}{\sysname}

\newcommand{\fsftOrig}{ICFT}
\newcommand{\ourNoCipher}{NoCipher}
\newcommand{\ourNoAnneal}{NoAnneal}
\newcommand{\ourNoSelect}{NoSelect}
\newcommand{\ICTL}{ICWL}
\renewcommand{\topfraction}{0.9}
\renewcommand{\bottomfraction}{0.8}
\renewcommand{\textfraction}{0.07}
\renewcommand{\floatpagefraction}{0.75}
\renewcommand{\dbltopfraction}{0.9}
\setcounter{dbltopnumber}{2}




\begin{abstract}
Fine-tuning LLMs with in-context examples (\ictrain) lets translation models adapt to new parallel data at deployment, but it is unclear whether they actually learn the translations of rare terms from those examples. We formalize this capability as in-context term learning (ICTL).
We show that standard \ictrain\ develops this capability only weakly, and propose \sysname, which strengthens it without any lexical resources. \sysname{} pairs each training target with in-context examples that share a rare term with it. It then replaces that term with a character-shuffled cipher in both the context and the reference, so the term can only be recovered by copying from context. The ciphered fraction of training data is annealed to preserve translation quality. Across seven translation directions between English and Tamil, Assamese, Mizo, Nepali and Hausa, and four models from 1B to 7B parameters, \sysname\ raises term recall over standard \ictrain\ by 5–15 absolute points without lowering COMET. 
Probes using synonyms, nonce words and controls for shared surface context show that these gains come from linking source terms to their in-context translations rather than from shortcuts.
\end{abstract}

\section{Introduction}
\label{sec:intro}
The in-context learning (ICL) capability of LLM-based machine translation systems has been found to be effective for continuous learning and adaptation, particularly for low resource languages where labeled data may not be available in one go~\citep{moslem-etal-2023-adaptive,li-etal-2025-context-learning-new,ramos-etal-2025-grammamt,zebaze-etal-2025-context}.
A complementary line of work has shown that this adaptation ability is further strengthened when the model is fine-tuned with in-context examples, a strategy we refer to as \ictrain~\citep{min2022metaicllearninglearncontext,alves-etal-2023-steering,zheng2024finetuninglargelanguagemodels,bornschein2025finetuned}.
While \ictrain{} improves overall translation quality via adaptation with new examples, a more fine-grained question has received less attention: does the model actually learn \emph{new or rare terms} from the in-context examples?
\begin{figure}[t]
\small
\definecolor{ctermc}{HTML}{1F5FAD}
\newcommand{\cterm}[1]{\textcolor{ctermc}{\underline{\textbf{#1}}}}
\newcommand{\exrow}[3][1.8em]{\raggedright\noindent\hangindent=#1\hangafter=1\makebox[#1][l]{\textbf{#2}}#3\par}
\newcommand{\runrow}[2]{\raggedright\noindent\hangindent=1em\hangafter=1\textbf{#1}~#2\par}
\newcommand{\outrow}[3]{\noindent\fcolorbox{#1}{#1}{\parbox{\dimexpr\columnwidth-2\fboxsep-2\fboxrule\relax}{\runrow{#2}{#3}}}\par}
\setlength{\fboxsep}{2pt}

\noindent\fbox{\parbox{\dimexpr\columnwidth-2\fboxsep-2\fboxrule\relax}{%
\textbf{In-context example}\par\vspace{1pt}
\exrow{$\vx^1$:}{\tatext{தேசிய அறிவியல் தினத்தையொட்டி, \cterm{விஞ்ஞானிகளுக்கு} \ldots}}
\exrow{$\vy^1$:}{\ldots{} has greeted the \cterm{scientists} on National Science Day.}}}

\vspace{2pt}

\noindent\fbox{\parbox{\dimexpr\columnwidth-2\fboxsep-2\fboxrule\relax}{%
\textbf{In-context example (distractor)}\par\vspace{1pt}
\exrow{$\vx^2$:}{\tatext{\ldots{} வெற்றி கரமாக ஏவிய \ldots}}
\exrow{$\vy^2$:}{\ldots{} the team for the successful \xterm{launch}.}}}

\vspace{2pt}

\noindent\fbox{\parbox{\dimexpr\columnwidth-2\fboxsep-2\fboxrule\relax}{%
\runrow{Test $\vxt$:}{\tatext{இந்த சாதனையை பாராட்டி பிரதமர் \cterm{விஞ்ஞானிகளுக்கு} தொலைபேசி மூலமாக வாழ்த்து தெரிவித்துள்ளார்.}}}}

\vspace{2pt}

\outrow{gray!12}{Gold $\vyt$:}{The Prime Minister applauded the \cterm{scientists}, calling their achievement an inspiring one.}
\vspace{1pt}
\outrow{green!8}{\method:}{The Prime Minister congratulated the \cterm{scientists} on the achievement through a phone call.~\cmark}
\vspace{1pt}
\outrow{red!8}{\fsftOrig:}{The Prime Minister congratulated the team for the successful \xterm{launch}.~\xmark}

\caption{Motivating example of in-context term learning (ICTL), Tamil$\to$English.
The test source $\vxt$ and $\vx^1$ share the term \protect\tatext{\cterm{விஞ்ஞானிகளுக்கு}} (``to the scientists''), so the model must find its translation \cterm{scientists} in $\vy^1$ and use it. The distractor $\vx^2$ shares the headline but is about a \xterm{launch}, which $\vxt$ never mentions. \fsftOrig{} copies $\vy^2$ verbatim, while \method{} copies \cterm{scientists}. Transliterations in Figure~\ref{fig:translit-ta}.}
\label{fig:example}
\end{figure}

Consider a test sentence $\vxt$ paired with an in-context example $(\vx^i, \vy^i)$.
If a rare source term $u$ appears in both $\vxt$ and $\vx^i$, and its translation $v$ appears in $\vy^i$, then the model has the opportunity to infer that $u$ translates to $v$ and place $v$ at the correct position in its output translation of $\vxt$ (See Figure~\ref{fig:example} for an example.)
We call this capability \emph{in-context term learning} (ICTL).
ICTL requires the model to perform a chain of implicit reasoning: to produce the rare term $v$, the model needs to locate its aligned word in the source $\vxt$, align $v$ in $\vy^i$ to corresponding $u$ in $\vx^i$, in-context lookup $v$ using the shared source word $u$,  and copy its translation to the appropriate output position.
We show that standard \ictrain{}, which we call In-Context Fine-Tuning (\fsftOrig), which simply maximizes likelihood of a target with similar examples in-context, does not sufficiently develop this chain.


Our goal in this paper is to strengthen such an implicit reasoning chain so that the model reliably extracts terms from in-context examples at inference time.  We propose \method{}, that promotes ICTL during \ictrain\ via two key ideas: First, we selectively include in-context examples that share rare terms with the target.  In \ictrain{}, the model can predict a target token through either
(i)~\emph{parametric recall} or (ii)~\emph{contextual copying}.
Parametric recall suffices for high-frequency terms but fails on rare, domain-specific vocabulary absent from the fine-tuning corpus.  By pairing examples with context that include shared rare terms, we are more likely to develop the ICTL capability.
%
Second, we reinforce the effect further via a cipher-based data augmentation that replaces selected rare target-language tokens with synthetic \emph{ciphers} obtained by  a character-level shuffling of the original term.  However, since ciphers distort original data distribution, we introduce a simple annealing to linearly decrease the fraction of ciphered training instances.

We evaluate \method{} on seven language pairs spanning diverse rare languages (Assamese, Tamil, Mizo, Nepali, Hausa).
%
We fine-tune four base models (Llama-3.2-1B, Qwen-2.5-1.5B, Qwen-2.5-3B, OLMo-3-7B).
Our experiments show that \method{} improves Term Recall over \fsftOrig{} on all models without deteriorating overall translation quality.
We present an analysis to provide evidence that our training does promote sound ICTL capabilities, while avoiding various learning short-cuts. 


In this work our goal was to analyze and promote terminology learning from raw in-context examples.  There is extensive prior work on promoting terminology translation in NMT, but they all require \emph{external lexical resources} in the form of either a synonym dictionary, or a bilingual term dictionary.  We compare with recent existing work that invoke an external alignment library to extract translation dictionary.  We show that \sysname\ better preserves fluency when incorporating term dictionaries than an existing state-of-the-art method (DragFT). 

\paragraph{Contributions.}
(1)~We formalize in-context term learning (ICTL) as a distinct capability, that allows a model to harness parallel in-context sentences to learn term translations.
(2)~We propose \method{} that strengthens ICTL without requiring external lexical resources, through a combination of how examples are paired with context, and via selective ciphering.
%
(3)~We demonstrate consistent improvements in terminology extraction across seven language pairs and three model families. \todo{Give some summary numbers.}
(4)~We analyze the models with various probes that show evidence that \method{}'s training promotes sound ICTL, without degenerating to learning shortcuts.
(5) We show that \method{} achieves higher COMET for comparable term-recall than methods like DragFT~\citep{zheng2024finetuninglargelanguagemodels}, that rely on an explicit bilingual term dictionary.






\section{Related Work}
\label{sec:related_work}

\paragraph{ICL for MT.}
ICL has proven effective for continuous adaptation in MT~\citep{moslem-etal-2023-adaptive}, with similarity-based example selection, especially for low-resource languages~\citep{zebaze-etal-2025-context}.  \citet{scoi2024} show that syntax- and coverage-aware retrieval of ICL examples improves adaptation,
\citet{li-etal-2025-context-learning-new} show that ICL with parallel examples outperforms dictionary-based prompting for low-resource translation, and \citet{ramos-etal-2025-grammamt} find that learning from examples is more effective than instruction tuning.  These observations motivate our focus on training models to extract terms from parallel sentences instead of glossaries.

\paragraph{\ictrain{} for MT.}
\citet{min2022metaicllearninglearncontext} first proposed fine-tuning with in-context examples to improve ICL, an idea since applied across diverse domains\citep{bornschein2025finetuned}  including logical reasoning~\citep{bertolazzi-etal-2026-teaching}, neuroscience~\citep{yu2025meta}, and log parsing~\citep{Yu_2026}.
%
%
For MT specifically, \citet{alves-etal-2023-steering} showed that \ictrain{} preserves ICL ability often lost during standard zero-shot fine-tuning.
\citet{malu2026contrastmattersunderstandingrobustness} show that \ictrain{} with random contexts weakens ICL, while similar contexts encourage label copying; their Contrastive-Context data recipe, adds contrast to make the model robust across the full similarity spectrum.
%
%
%
Other work fine-tunes per domain with varying data sizes~\citep{vieira-etal-2024-much}, preference pairs~\citep{vieira2025data}, multi-stage schedules for shifting domains~\citep{guan-etal-2025-multi-new}, or meta-learning with domain-specific data~\citep{sharaf-etal-2020-meta}; our goal instead is a single model that adapts to new domains via ICL at inference time.

\paragraph{Terminology-aware MT.}
Prior work enforces terminology through constrained decoding~\citep{chatterjee-etal-2022-accurate,chen-etal-2021-lexically} or glossary-augmented LLM prompting~\citep{pmlr-v202-zhang23m, terminology_prompting_2023,domain_terminology_llm_2023,ghazvininejad2023dictionarybasedphraselevelpromptinglarge} 
.
\citet{zheng2024finetuninglargelanguagemodels}'s DragFT integrates bilingual glossary entries into source sentences to improve domain terminology.
Unlike these methods, we assume no lexical resources at inference or training time: neither bilingual glossaries of term translations nor monolingual thesauri of target-language synonyms.  Both are scarce for low-resource languages.  Most prior work on terminology learning are on relatively high resource languages. 
Instead, we train the model to \emph{infer} terminology mappings from parallel in-context demonstrations.  The prompting tricks used in the TL literature are orthogonal to our focus in this paper of  enhancing in-context term learning.
%
\citet{mutal-etal-2025-factors} find that one to three in-context examples matching the test input's domain and register give the largest gains in medical MT, but they score only whole sentences and don't check whether terms are carried over from demonstrations. 
%


\paragraph{Ciphers and synthetic tokens.}
\citet{fang-etal-2025-icl} use substitution ciphers as a diagnostic to disentangle task retrieval from task learning in ICL, showing that LLMs can decode latent bijective mappings from in-context demonstrations. \citet{agarwal2025} show with synthetic sequence to sequence tasks that, LLMs struggle to uncover input-output alignments during ICL, and propose instead \ictrain\ to teach the missing alignment knowledge.
%
\citet{bafna2026rashid} use ciphers to create synthetic low resource languages where they replace all tokens for a new task.
Our work draws on a related intuition but uses cipher tokens not as a diagnostic probe but as a \emph{training signal} that forces the model to develop robust contextual-copying circuits for cross-lingual terminology extraction.
To our knowledge, this is the first application of cipher-based data augmentation to explicitly strengthen in-context term learning in MT.

\section{Problem Formulation}
We are given a corpus $\mathcal{D}$ of parallel sentence pairs for translation between two languages referred to as source and target languages.  Our goal is to fine-tune an LLM $\model$ to teach the translation task while maintaining the ability to learn from new labeled example pairs $\Ddom$ that become available during deployment. Specifically, the trained model should be able to harness $\Ddom$ to learn to translate rare (in $D$) source terms based purely on their usage in parallel in-context examples.  
We assume that no term-level supervision is given at any stage, $\Ddom$ contains only sentence pairs, with no glossary or alignment.


Since our goal is to fine-tune $\model$ to support such in-context learning, instead of standard fine-tuning with independently sampled ($\vx,\vy)$ pairs from $D$, we fine-tune with in-context examples.
Thus, while standard fine-tuning maximizes $\mathbb{E}_{(\vxt,\vyt) \sim \mathcal{D}} 
\sum_t\log P_\theta(y^*_t \mid \vxt, \vyt_{<t})$, \ictrain\ instead samples for every training example $(\vxt,\vyt)$, 
$k$ examples similar to $\vxt$ from the labeled pool $D$ to be used as in-context examples.  Let $C = \{(\vx^1,\vy^1), \ldots, (\vx^k,\vy^k)\}$ denote the $k$ parallel examples with the highest similarity to $\vxt$.  During training, the input prompt comprises $C,\vxt$, and the training objective maximizes  the likelihood of the target $\vyt$:
\begin{equation}
\begin{split}
\sum_t\log P_\theta\left(y^*_t \mid \vx^1,\vy^1, \ldots, \vx^k,\vy^k, \vxt, \vyt_{<t}\right)
\end{split}
\label{eq:ictrain}
\end{equation}
We call this method \ictrain.
As shown in previous work, and also validated in our experiments, fine-tuning a translation model in the \ictrain\ setting significantly enhances the ease and accuracy of in-context learning from limited labeled examples that become available at test-time. 

In this paper, we go beyond simply measuring aggregate accuracy gains, and take a more detailed look at how well the fine-tuned model learns to harness rare terms from the context examples as shown in Figure~\ref{fig:example}.



\section{\sysname}
\label{sec:data_aug_cipher}
%
We start with the observation that, under the \ictrain{} objective (Eq.~\eqref{eq:ictrain}), the model can predict a target token $\yt_t$ through two distinct mechanisms:
(i)~\emph{Parametric recall}, where the feed-forward layers retrieve a memorized translation of the source token.  Developing this memory requires high frequency of the term during training, and is likely to fail on rare terms absent from the fine-tuning corpus.
(ii)~\emph{Contextual copying}, where the attention layers couple words in $\vy$ to their aligned word in the corresponding $\vx$, and use the shared source term between $\vxt$ and an in-context source sentence $\vx^i$ to copy its translation to the output. 
Our goal is therefore to strengthen the contextual-copying pathway so that the model reliably extracts terminology from in-context examples at inference time.
\label{sec:alignment_and_copying}

%
Let $\alignT{\vx}{t}$ denote the source token in sentence $\vx$ that is aligned to the $t$-th token of its translation $\vy$; that is, $\alignT{\vx}{t}$ is the source word responsible for generating $y_t$.
Consider a target token $\yt_t$ in the gold translation $\vyt$ of the test source $\vxt$, and suppose the same underlying term appears at position $s$ in an in-context target $\vy^i$.  Let $u$ denote their common source-language form and $v$ their common target-language form, so that:
\begin{equation}
\label{eq:copy_condition}
    y^i_s = \yt_t = v,
    \qquad
    \alignT{\vx^i}{s} = \alignT{\vxt}{t} = u.
\end{equation}
%
When condition~\eqref{eq:copy_condition} holds, the model can, in principle, predict $\yt_t$ by (a)~at decoding step $t$, inferring $A(t)$ and identifying $\alignT{\vxt}{t}$ as $u$, (b)~aligning words in the examples $\vy^i$ to their corresponding translation in the source, in particular  $v = y^i_s$ to $\alignT{\vx^i}{s}$ and copying it to form an aligned induction circuit~\cite{olsson2022incontextlearninginductionheads}, (c)~looking up context $\vy^i$ tokens with key as $\alignT{\vxt}{t}$, to retrieve $y^i_s$ based on match with $\alignT{\vx^i}{s}$, (d)~copying $v$ to position $t$ of the output.

Training the model to carry out this chain reliably is the central objective of our method. However, we have to do so without assuming availability of term-level alignment between sentences during training.

\xhdr{Overview of \sysname}
Our key idea is to pick context-target pairs such that there exists a shared rare word between $\vy^i$ and $\vyt$.  Additionally, for rare shared words, parametric learning will likely fail, and the model will be forced to copy from the context.   Assuming well-formed translation pairs,  $\vx^i$ and $\vxt$ are likely to share their respective aligned source word, which provides the pathway for contextual copying.  However, given limited training data in low resource languages, there may not be enough such examples of shared rare words, and the model may quickly overfit.  So, we dynamically replace the shared words in $\vy^i$ and $\vyt$ with ciphers. Section~\ref{sec:cipher_replacement} describes this in detail.  Training with synthetic ciphers tokens could encourage ICTL but in the process interfere with the normal translation task.  We provide details of how we use an annealing schedule to counter such interference in Section~\ref{sec:anneal}.
Figure~\ref{fig:overview} presents an overview of our method.

\definecolor{cfSharedLine}{HTML}{2A78D6}
\definecolor{cfSharedOrig}{HTML}{CDE2FB}
\definecolor{cfSharedCip}{HTML}{86B6EF}
\definecolor{cfDistLine}{HTML}{EB6834}
\definecolor{cfDistOrig}{HTML}{FCD9C4}
\definecolor{cfDistCip}{HTML}{F4A582}
\definecolor{cfFrame}{HTML}{898781}
\definecolor{cfArrow}{HTML}{52514E}
\definecolor{cfPanelA}{HTML}{EFF6EC}\definecolor{cfPanelAEdge}{HTML}{7FB069}
\definecolor{cfPanelB}{HTML}{EEF3FA}\definecolor{cfPanelBEdge}{HTML}{6C9BD2}
\definecolor{cfPanelC}{HTML}{FCF4E4}\definecolor{cfPanelCEdge}{HTML}{E0A84A}
\definecolor{cfAlgo}{HTML}{D9E7F7}\definecolor{cfAlgoEdge}{HTML}{4F7FBF}
\definecolor{cfPool}{HTML}{D8ECD2}\definecolor{cfPoolEdge}{HTML}{6AA05A}
\definecolor{cfIndex}{HTML}{E6DDF1}\definecolor{cfIndexEdge}{HTML}{8A6FB5}
\definecolor{cfLLM}{HTML}{F6D7A7}\definecolor{cfLLMEdge}{HTML}{C98A2C}

\newsavebox{\cfspanbox}
\DeclareRobustCommand{\cfspan}[3]{%
  \begingroup
  \sbox{\cfspanbox}{\hspace{0.6pt}#3\hspace{0.6pt}}%
  \setlength{\fboxsep}{0pt}%
  \colorbox{#1}{\rule[-0.3em]{0pt}{1.0em}\usebox{\cfspanbox}}%
  \llap{\textcolor{#2}{\rule[-0.3em]{\wd\cfspanbox}{0.9pt}}}%
  \endgroup}
\DeclareRobustCommand{\cfShared}[1]{\cfspan{cfSharedOrig}{cfSharedLine}{#1}}
\DeclareRobustCommand{\cfSharedCip}[1]{\cfspan{cfSharedCip}{cfSharedLine}{#1}}
\DeclareRobustCommand{\cfDist}[1]{\cfspan{cfDistOrig}{cfDistLine}{#1}}
\DeclareRobustCommand{\cfDistCip}[1]{\cfspan{cfDistCip}{cfDistLine}{#1}}

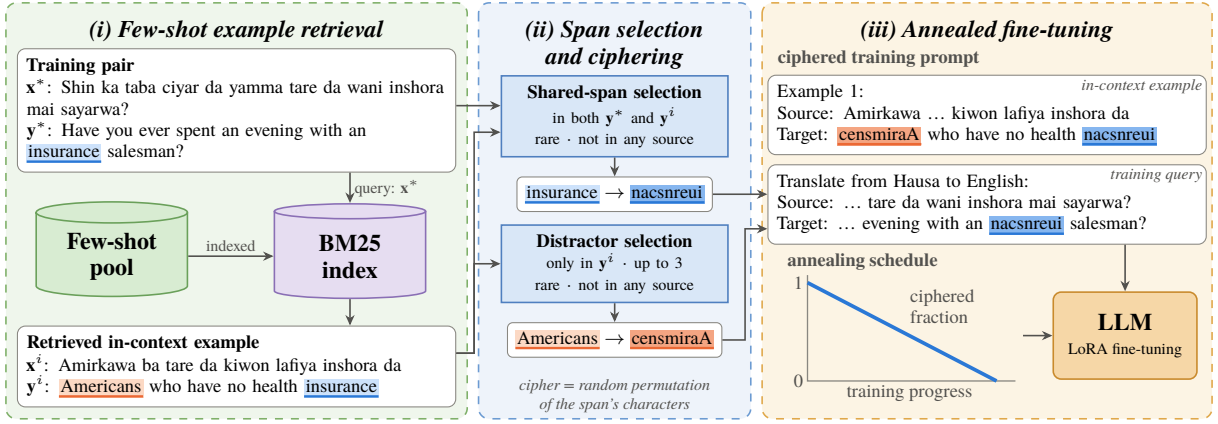
\begin{figure*}[t]
\centering
\begin{tikzpicture}[
    font=\scriptsize,
    panel/.style={draw=#1Edge, fill=#1, dashed, line width=0.6pt, rounded corners=4pt},
    ptitle/.style={font=\footnotesize\bfseries\itshape, anchor=north, align=center},
    sent/.style={draw=cfFrame, fill=white, line width=0.4pt, rounded corners=3pt,
                 align=left, inner sep=3.5pt,
                 execute at begin node={\hyphenpenalty=10000\exhyphenpenalty=10000}},
    algo/.style={draw=cfAlgoEdge, fill=cfAlgo, line width=0.6pt, align=center,
                 inner sep=3.5pt, text width=2.75cm},
    chip/.style={draw=cfFrame, fill=white, line width=0.4pt, rounded corners=2pt,
                 inner sep=3pt},
    db/.style={cylinder, shape border rotate=90, aspect=0.22, draw=#1Edge, fill=#1,
               line width=0.6pt, minimum width=2.0cm, minimum height=1.25cm,
               align=center, inner sep=1pt, font=\small\bfseries},
    tag/.style={font=\tiny\itshape, text=cfArrow, inner sep=1.5pt, anchor=north east},
    arr/.style={-{Stealth[length=4pt,width=3.5pt]}, line width=0.6pt, draw=cfArrow},
    lab/.style={font=\tiny, text=cfArrow, inner sep=1.5pt},
]

\node[sent, text width=5.55cm, anchor=north] (pair) at (3.05,-0.6) {%
    \textbf{Training pair}\\[0.5pt]
    $\vxt$: Shin ka taba ciyar da yamma tare da wani inshora mai sayarwa?\\
    $\vyt$: Have you ever spent an evening with an \cfShared{insurance} salesman?};

\node[db=cfPool]  (pool) at (1.4,-3.35) {Few-shot\\pool};
\node[db=cfIndex] (bm25) at (4.55,-3.35) {BM25\\index};
\draw[arr] (pool) -- node[lab, above] {indexed} (bm25);
\draw[arr] (pair.south -| bm25) -- node[lab, right] {query: $\vxt$} (bm25.north);

\node[sent, text width=5.55cm, anchor=south] (ex) at (3.05,-5.35) {%
    \textbf{Retrieved in-context example}\\[0.5pt]
    $\vx^i$: Amirkawa ba tare da kiwon lafiya inshora~da\\
    $\vy^i$: \cfDist{Americans} who have no health \cfShared{insurance}};
\draw[arr] (bm25.south) -- (ex.north -| bm25);

\node[algo] (fshared) at (8.05,-1.5) {%
    \textbf{Shared-span selection}\\[1pt]
    {\tiny in both $\vyt$ and $\vy^i$\\ rare $\cdot$ not in any source}};
\node[chip, below=2.5mm of fshared] (cshared)
    {\cfShared{insurance} $\rightarrow$ \cfSharedCip{nacsnreui}};
\draw[arr] (fshared) -- (cshared);

\node[algo] (fdist) at (8.05,-3.45) {%
    \textbf{Distractor selection}\\[1pt]
    {\tiny only in $\vy^i$ $\cdot$ up to 3\\ rare $\cdot$ not in any source}};
\node[chip, below=2.5mm of fdist] (cdist)
    {\cfDist{Americans} $\rightarrow$ \cfDistCip{censmiraA}};
\draw[arr] (fdist) -- (cdist);

\node[font=\tiny\itshape, text=cfArrow, align=center, below=1.5mm of cdist]
    {cipher = random permutation\\of the span's characters};

\coordinate (fsin) at ([yshift=4pt]fshared.west);
\draw[arr] (fsin -| pair.east) -- (fsin);
\draw[arr] ([yshift=5pt]ex.east) -| ([xshift=-0.35cm, yshift=-6pt]fshared.west) -- ([yshift=-6pt]fshared.west);
\draw[arr] ([yshift=5pt]ex.east) -| ([xshift=-0.35cm]fdist.west) -- (fdist.west);

\node[font=\scriptsize\bfseries, text=cfArrow, anchor=north west, inner sep=0pt]
    at (10.2,-0.62) {ciphered training prompt};
\node[sent, text width=5.55cm, anchor=north] (icl) at (12.975,-0.95) {%
    Example 1:\\
    Source: Amirkawa \ldots\ kiwon lafiya inshora~da\\
    Target: \cfDistCip{censmiraA} who have no health \cfSharedCip{nacsnreui}};
\node[tag] at (icl.north east) {in-context example};
\node[sent, text width=5.55cm, anchor=north] (query) at ([yshift=-1.2mm]icl.south) {%
    Translate from Hausa to English:\\
    Source: \ldots\ tare da wani inshora mai sayarwa?\\
    Target: \ldots\ evening with an \cfSharedCip{nacsnreui} salesman?};
\node[tag] at (query.north east) {training query};

\draw[arr] (cshared.east) -- (cshared.east -| query.west);
\draw[arr] (cdist.east) -- ++(0.3,0) |- ([yshift=6pt]query.south west);

\begin{scope}[shift={(10.6,-5.0)}, lab/.append style={font=\scriptsize}]
    \draw[cfFrame, line width=0.5pt] (0,1.4) -- (0,0) -- (2.75,0);
    \draw[cfSharedLine, line width=1.4pt] (0,1.3) -- (2.5,0);
    \node[lab, anchor=east] at (0,1.3) {1};
    \node[lab, anchor=east] at (0,0) {0};
    \node[lab, anchor=north] at (1.35,0) {training progress};
    \node[lab, anchor=south west, align=left] at (1.3,0.7) {ciphered\\fraction};
\end{scope}

\node[draw=cfLLMEdge, fill=cfLLM, line width=0.6pt, rounded corners=3pt,
      minimum width=1.9cm, minimum height=1.15cm, align=center] (llm) at (14.8,-4.4)
    {\footnotesize\textbf{LLM}\\[1pt]{\tiny LoRA fine-tuning}};
\draw[arr] (query.south -| llm) -- (llm.north);
\draw[arr] (13.45,-4.4) -- (llm.west);
\node[font=\scriptsize\bfseries, text=cfArrow, anchor=west] at (10.2,-3.42) {annealing schedule};

\begin{scope}[on background layer]
    \draw[panel=cfPanelA] (0,0) rectangle (6.1,-5.5);
    \draw[panel=cfPanelB] (6.25,0) rectangle (9.85,-5.5);
    \draw[panel=cfPanelC] (10.0,0) rectangle (15.95,-5.5);
\end{scope}
\node[ptitle] at (3.05,-0.1) {(i) Few-shot example retrieval};
\node[ptitle] at (8.05,-0.1)  {(ii) Span selection\\and ciphering};
\node[ptitle] at (12.975,-0.1) {(iii) Annealed fine-tuning};

\end{tikzpicture}
\caption{\label{fig:overview} Overview of \method.
\emph{(i)}~In-context examples are retrieved from the few-shot pool with BM25.
\emph{(ii)}~A \cfShared{shared rare span}, occurring in both the gold reference $\vyt$ and
the ICL target $\vy^i$, is replaced by its \cfSharedCip{cipher}, a random permutation of
its characters, at every occurrence in the prompt and the reference; \cfDist{distractor spans},
occurring only in $\vy^i$, are likewise replaced by their \cfDistCip{ciphers}.
\emph{(iii)}~Training is annealed linearly from ciphered to original instances.}
\end{figure*}

\subsection{Training Data Construction}
\label{sec:cipher_replacement}
We first identify target-language shared word spans, a sequence of consecutive space-separated words, that occur in both the gold reference $\vyt$ and at least one in-context target sentence $\vy^i$ within the same training prompt.  
These spans mark positions where the condition in Eq~\eqref{eq:copy_condition} is satisfied and where cipher replacement can meaningfully encourage contextual copying.
A candidate span is retained only if it satisfies the following criteria:
\begin{enumerate}[leftmargin=*]

    \item \textbf{Lexical rarity.}
    Every word in the span occurs at most~$\beta$ times in the training pool.
    %
    As shown in previous studies~\cite{Chan2022,Chan2024TowardUI} rare terms lead to development of in-context learning pathways over parametric learning.   

    \item \textbf{Source exclusion.}
    The span does not appear in any source sentence.
    This ensures that we do not interfere with developing the skill to learn to copy occasional tokens (typically digits and abbreviations) from the source sentence as needed. Instead, we reinforce cases where the model must use cross-lingual alignment through the \emph{target} sentence to recover the term.
\end{enumerate}

\noindent
For each shared span, we replace it with a \emph{cipher}.
The cipher for a span is a random permutation of its characters, taken after removing whitespaces and any punctuation at its start and end. Ciphering promotes the model to learn to refer to the in-context examples for the translation of rare terms since the ciphers are unrecognizable by the model, and it cannot rely on its parametric knowledge to recover them.
%
Every occurrence of the span in $\vyt$ and in all matching in-context targets $\vy^i$ is replaced by the same cipher. For each instance, only one shared span and cipher are chosen. 
Along with this at most three distractor spans, which satisfy the conditions above but do not occur in the gold reference $\vyt$, are chosen and replaced with ciphers as shown in Figure~\ref{fig:overview}.  Such distractors serve the purpose of preventing the model from learning any shortcuts arising out of ciphering.

\subsection{Annealed Training with Ciphered Text}
\label{sec:anneal}
While training on ciphered text, could promote contextual-copying, it may interfere with normal training since ciphered words are not part of the target language. We therefore anneal the training data linearly from pure cipher data to pure original data so that early on the model learns the correct mechanism and on later parts the model would learn how to handle real test data without ciphers.

Let $t \in \{1, \dots, T\}$ index the training steps, and $\tau_t = (t-1)/(T-1)$ be the training progress at step $t$.
Each instance $n$ draws $u_n \sim \mathcal{U}[0,1)$ once. 
The ciphered version $\cvyt$ is used for training if $u_n \geq \tau_t$, else the original version $\vyt$ is used.
%
%
The ciphered fraction thus falls linearly from one to zero, and the run costs about the same as \fsftOrig.

\section{Empirical Analysis and Evaluation}
\label{sec:experimental-setup}
\vspace{-0.07in}
We present the results of an extensive empirical study designed to answer the following questions:
\begin{enumerate}[label=\textbf{RQ\arabic*}, ref=RQ\arabic*, leftmargin=*]
    
    \item\label{rq:1} Can \sysname's synthetic ciphering help promote term recall without adversely affecting normal translation quality?
    \item\label{rq:2} What in the composition of context-target pairs help promote ICTL in \sysname?
    \item\label{rq:3} Could \sysname\ be using learning shortcuts to pick rare terms from the context?
    \item\label{rq:5} Can \sysname\ directly harness term translations from raw parallel sentences in context, at par or better than existing methods using explicit term dictionaries?
\end{enumerate}
We answer these questions with experiments on seven translation tasks over four models:

\xhdr{Datasets}
\label{sec:datasets}
We experiment on 7 machine translation (MT) tasks from Tamil, Assamese, Mizo, Nepali, Hausa to English and English to Assamese, Hausa. The datasets used for each language pair and their details are listed in Table~\ref{tab:dataset-mt} (Appendix~\ref{app:datasets}). 
\xhdr{Few-Shot ICL Dataset Construction}
For each language pair, we construct training instances with in-context learning (ICL) examples. We partition the training data $D$ into training, validation, and the few-shot retrieval pool.
For every training example, the number of ICL examples $k$ is sampled uniformly from $\{1,\dots,5\}$ to ensure that it won't overfit to a fixed number of examples. Demonstrations are retrieved using BM25 similarity over source sentences tokenized with IndicBERT \cite{doddapaneni2023leavingindiclanguagebehind}
for Tamil, Assamese and Nepali, and with SentencePiece \cite{kudo2018sentencepiecesimplelanguageindependent} for English, Mizo and Hausa. Each instance is serialized into a prompt template containing the selected $(\text{source}, \text{target})$ demonstration pairs and  the source sentence.
An example prompt for Mizo$\to$English is shown in Figure~\ref{box:anecdote2}.
For testing, we fix $k$ to 5, and all the $k$ examples are retrieved using BM25 similarity.

\xhdr{Models and Fine-tuning details}
We experiment with the following open-source LLMs:
\begin{enumerate*}[(1)]
    \item Llama-3.2-1B~\cite{llama3}
    \item Qwen-2.5-1.5B~\cite{qwen2025qwen25technicalreport}
    \item Qwen-2.5-3B~\cite{qwen2025qwen25technicalreport}
    \item OLMo-3-7B~\cite{olmo2026olmo3}
\end{enumerate*}. These models are LoRA fine-tuned. More details in Appendix~\ref{sec:app:ft}.

For a language pair, across all training variants, we have used the same $\vxt$ for training, but the ICL examples are different for each variant.

\xhdr{Evaluation Metrics}
We evaluate translation quality using BLEU \cite{post-2018-call} and COMET \cite{rei-etal-2022-comet}. We use \textit{Term Recall} to score how many rare terms are picked up in the prediction, similar to what is defined in \citet{semenov-etal-2025-findings,alam2021evaluationmachinetranslationterminology}.  We create a pool of candidate terms $C_i$ that are rare in the training data and shared between context and target (More details in Appendix~\ref{sec:app:tr}).
Let $\hat{\vy}^i$ denote the generated translation for evaluation instance $i$. Term recall is computed as the fraction of candidate terms that appear in the generated output:
$
\mathrm{Recall}(i)
=
\frac{\sum_{t \in C_i} \mathbb{I}[t \in \hat{\vy}^i]}{|C_i|}
$
where $\mathbb{I}$ is the indicator function.
As in \citet{semenov-etal-2025-findings}, we lemmatize all words in $\hat{\vy}^i$ and $C_i$ before computing term recall.
%



\subsection{Methods Compared}
\label{sec:methods}
In this paper we will compare with these methods:
\begin{enumerate}[leftmargin=*,itemsep=0pt]
    \item \textbf{Base}: The model without any fine-tuning.
    \item \textbf{\fsftOrig}:\ Here the training data is selected without the conditions in Section~\ref{sec:cipher_replacement}. So, it is not compulsory to have shared word spans, and no ciphering is applied here.
    \item \textbf{\sysname}: Our proposed method, with $\beta$, as described in Section~\ref{sec:cipher_replacement}, set to 250. We also perform an ablation study on $\beta$ in Section~\ref{sec:analysis:rq2}.
    \item \textbf{DragFT}~\cite{zheng2024finetuninglargelanguagemodels}:
    Although, our setup and focus in this paper is different from the extensive literature on MT with glossary-based term learning (GlossTL), we present a separate comparison with a state-of-the-art GlossTL method, DragFT.  Given a glossary $G$, this method proposes to replace matching source words in $\vxt$ by their target translations in $G$.  For training, such a model they require using a global dictionary, and we built this by running an alignment method eflomal~\cite{Ostling2016efmaral} on the training and testing few-shot pools.  DragFT also incorporates examples in the context, but those are meant for improving overall translation, and not term learning.  We present this comparison in Section~\ref{sec:analysis:rq5}.
\end{enumerate}

\section{Results and Analysis}
\label{sec:rq-answers}
\begin{table*}[t]
\centering
\small
\setlength{\tabcolsep}{3pt}
\renewcommand{\arraystretch}{1.0}
\setlength{\aboverulesep}{0pt}
\setlength{\belowrulesep}{0pt}
\setlength{\extrarowheight}{0pt}
\resizebox{\textwidth}{!}{%
\begin{tabular}{llc>{\columncolor{gray!12}}cc>{\columncolor{gray!12}}cc>{\columncolor{gray!12}}cc>{\columncolor{gray!12}}ccc}
\toprule
\rule{0pt}{2.3ex}Pair & Training & \multicolumn{2}{c}{Llama-3.2-1B} & \multicolumn{2}{c}{Qwen2.5-1.5B} & \multicolumn{2}{c}{Qwen2.5-3B} & \multicolumn{2}{c}{OLMo-3-7B} & \multicolumn{2}{c}{Support} \\
\cmidrule(lr){3-4} \cmidrule(lr){5-6} \cmidrule(lr){7-8} \cmidrule(lr){9-10} \cmidrule(lr){11-12}
 &  & COMET & TermRecall & COMET & TermRecall & COMET & TermRecall & COMET & TermRecall & COMET & TermRecall \\
\midrule
\multirow{3}{*}{Lus-En} & \rule{0pt}{2.3ex}Base & 45.56 & 54.68 & 45.57 & 55.53 & 46.88 & 54.26 & 46.40 & 50.43 & \multirow{3}{*}{2000} & \multirow{3}{*}{235} \\
 & \fsftOrig & 54.85 {\scriptsize(0.46)} & 53.69 {\scriptsize(2.56)} & 56.37 {\scriptsize(0.29)} & 59.93 {\scriptsize(1.81)} & 58.28 {\scriptsize(0.02)} & 62.70 {\scriptsize(1.09)} & 59.68 {\scriptsize(0.12)} & 69.48 {\scriptsize(1.78)} &  &  \\
 & \sysname & \cellcolor{green!25}\textbf{55.08} {\scriptsize(0.39)} & \cellcolor{green!68!gray!37}\textbf{61.63} {\scriptsize(1.07)} & \cellcolor{green!25}\textbf{56.46} {\scriptsize(0.86)} & \cellcolor{green!68!gray!37}\textbf{65.93} {\scriptsize(1.74)} & \cellcolor{green!25}\textbf{58.43} {\scriptsize(0.44)} & \cellcolor{green!68!gray!37}\textbf{68.65} {\scriptsize(0.54)} & \cellcolor{green!25}\textbf{60.23} {\scriptsize(0.38)} & \cellcolor{green!68!gray!37}\textbf{77.73} {\scriptsize(0.54)} &  &  \\
\midrule
\multirow{3}{*}{As-En} & \rule{0pt}{2.3ex}Base & 64.44 & 66.55 & 63.74 & 57.92 & 69.72 & 72.39 & 65.75 & 71.70 & \multirow{3}{*}{2000} & \multirow{3}{*}{568} \\
 & \fsftOrig & 68.53 {\scriptsize(0.20)} & 69.68 {\scriptsize(0.64)} & 67.92 {\scriptsize(0.14)} & 67.75 {\scriptsize(0.95)} & 71.48 {\scriptsize(0.12)} & 73.55 {\scriptsize(1.77)} & 71.38 {\scriptsize(0.08)} & 77.85 {\scriptsize(0.77)} &  &  \\
 & \sysname & \cellcolor{green!25}\textbf{69.15} {\scriptsize(0.22)} & \cellcolor{green!68!gray!37}\textbf{76.00} {\scriptsize(1.82)} & \cellcolor{green!25}\textbf{68.92} {\scriptsize(0.10)} & \cellcolor{green!68!gray!37}\textbf{75.39} {\scriptsize(1.41)} & \cellcolor{green!25}\textbf{71.92} {\scriptsize(0.21)} & \cellcolor{green!68!gray!37}\textbf{80.46} {\scriptsize(1.01)} & \cellcolor{green!25}\textbf{72.26} {\scriptsize(0.18)} & \cellcolor{green!68!gray!37}\textbf{83.36} {\scriptsize(0.91)} &  &  \\
\midrule
\multirow{3}{*}{Ta-En} & \rule{0pt}{2.3ex}Base & 62.21 & 36.97 & 56.92 & 33.31 & 64.59 & 46.71 & 69.60 & 58.90 & \multirow{3}{*}{1998} & \multirow{3}{*}{415} \\
 & \fsftOrig & 68.14 {\scriptsize(0.12)} & 45.75 {\scriptsize(0.35)} & 67.36 {\scriptsize(0.26)} & 42.06 {\scriptsize(1.98)} & 70.44 {\scriptsize(0.16)} & 54.26 {\scriptsize(1.58)} & 72.99 {\scriptsize(0.17)} & 59.93 {\scriptsize(1.84)} &  &  \\
 & \sysname & \cellcolor{green!25}\textbf{70.12} {\scriptsize(0.17)} & \cellcolor{green!68!gray!37}\textbf{57.75} {\scriptsize(0.15)} & \cellcolor{green!25}\textbf{69.41} {\scriptsize(0.20)} & \cellcolor{green!68!gray!37}\textbf{56.89} {\scriptsize(0.53)} & \cellcolor{green!25}\textbf{72.20} {\scriptsize(0.07)} & \cellcolor{green!68!gray!37}\textbf{65.11} {\scriptsize(0.76)} & \cellcolor{green!25}\textbf{74.40} {\scriptsize(0.13)} & \cellcolor{green!68!gray!37}\textbf{70.90} {\scriptsize(1.53)} &  &  \\
\midrule
\multirow{3}{*}{Ne-En} & \rule{0pt}{2.3ex}Base & 77.80 & 65.97 & 71.08 & 51.83 & 78.33 & 68.14 & 81.13 & \cellcolor{green!68!gray!37}\textbf{75.07} & \multirow{3}{*}{500} & \multirow{3}{*}{230} \\
 & \fsftOrig & 81.52 {\scriptsize(0.20)} & 60.55 {\scriptsize(3.36)} & 79.97 {\scriptsize(0.09)} & 58.55 {\scriptsize(1.62)} & 82.56 {\scriptsize(0.27)} & 66.37 {\scriptsize(1.52)} & 83.63 {\scriptsize(0.13)} & 69.16 {\scriptsize(1.16)} &  &  \\
 & \sysname & \cellcolor{green!25}\textbf{82.23} {\scriptsize(0.06)} & \cellcolor{green!68!gray!37}\textbf{66.70} {\scriptsize(2.14)} & \cellcolor{green!25}\textbf{81.00} {\scriptsize(0.40)} & \cellcolor{green!68!gray!37}\textbf{65.74} {\scriptsize(2.55)} & \cellcolor{green!25}\textbf{83.45} {\scriptsize(0.18)} & \cellcolor{green!68!gray!37}\textbf{71.48} {\scriptsize(0.86)} & \cellcolor{green!25}\textbf{84.15} {\scriptsize(0.05)} & 74.10 {\scriptsize(2.79)} &  &  \\
\midrule
\multirow{3}{*}{Ha-En} & \rule{0pt}{2.3ex}Base & 56.66 & 34.59 & 53.57 & 39.43 & 55.83 & 43.10 & 57.83 & 42.92 & \multirow{3}{*}{1997} & \multirow{3}{*}{186} \\
 & \fsftOrig & 66.64 {\scriptsize(0.12)} & 49.55 {\scriptsize(1.23)} & 66.56 {\scriptsize(0.29)} & 48.27 {\scriptsize(1.35)} & 67.92 {\scriptsize(0.18)} & 49.10 {\scriptsize(0.24)} & 69.80 {\scriptsize(0.18)} & 56.06 {\scriptsize(1.93)} &  &  \\
 & \sysname & \cellcolor{green!25}\textbf{68.97} {\scriptsize(0.07)} & \cellcolor{green!68!gray!37}\textbf{59.35} {\scriptsize(1.73)} & \cellcolor{green!25}\textbf{68.90} {\scriptsize(0.11)} & \cellcolor{green!68!gray!37}\textbf{58.39} {\scriptsize(2.21)} & \cellcolor{green!25}\textbf{70.66} {\scriptsize(0.14)} & \cellcolor{green!68!gray!37}\textbf{60.16} {\scriptsize(4.03)} & \cellcolor{green!25}\textbf{71.88} {\scriptsize(0.27)} & \cellcolor{green!68!gray!37}\textbf{63.29} {\scriptsize(1.75)} &  &  \\
\midrule
\multirow{3}{*}{En-As} & \rule{0pt}{2.3ex}Base & 52.04 & 34.85 & 46.88 & 31.38 & 52.17 & 40.84 & 54.76 & 39.86 & \multirow{3}{*}{2000} & \multirow{3}{*}{1288} \\
 & \fsftOrig & 69.59 {\scriptsize(0.02)} & 48.29 {\scriptsize(0.70)} & \cellcolor{green!25}\textbf{69.50} {\scriptsize(0.44)} & 47.30 {\scriptsize(0.47)} & 71.93 {\scriptsize(0.40)} & 50.17 {\scriptsize(1.35)} & 72.76 {\scriptsize(0.07)} & 54.82 {\scriptsize(0.88)} &  &  \\
 & \sysname & \cellcolor{green!25}\textbf{69.89} {\scriptsize(0.12)} & \cellcolor{green!68!gray!37}\textbf{53.56} {\scriptsize(0.88)} & 69.44 {\scriptsize(0.58)} & \cellcolor{green!68!gray!37}\textbf{52.10} {\scriptsize(0.81)} & \cellcolor{green!25}\textbf{72.43} {\scriptsize(0.19)} & \cellcolor{green!68!gray!37}\textbf{55.13} {\scriptsize(0.87)} & \cellcolor{green!25}\textbf{73.08} {\scriptsize(0.25)} & \cellcolor{green!68!gray!37}\textbf{60.65} {\scriptsize(1.27)} &  &  \\
\midrule
\multirow{3}{*}{En-Ha} & \rule{0pt}{2.3ex}Base & 43.97 & 19.68 & 46.00 & 23.16 & 43.49 & 22.66 & 46.02 & 23.96 & \multirow{3}{*}{1997} & \multirow{3}{*}{1339} \\
 & \fsftOrig & 73.88 {\scriptsize(0.18)} & 46.95 {\scriptsize(0.95)} & 73.10 {\scriptsize(0.07)} & 47.29 {\scriptsize(0.98)} & 74.44 {\scriptsize(0.08)} & 50.03 {\scriptsize(0.56)} & 74.78 {\scriptsize(0.19)} & 51.70 {\scriptsize(0.20)} &  &  \\
 & \sysname & \cellcolor{green!25}\textbf{75.99} {\scriptsize(0.20)} & \cellcolor{green!68!gray!37}\textbf{54.85} {\scriptsize(1.17)} & \cellcolor{green!25}\textbf{75.96} {\scriptsize(0.05)} & \cellcolor{green!68!gray!37}\textbf{54.70} {\scriptsize(0.07)} & \cellcolor{green!25}\textbf{76.82} {\scriptsize(0.18)} & \cellcolor{green!68!gray!37}\textbf{56.36} {\scriptsize(0.88)} & \cellcolor{green!25}\textbf{76.99} {\scriptsize(0.15)} & \cellcolor{green!68!gray!37}\textbf{57.32} {\scriptsize(0.30)} &  &  \\
\bottomrule
\end{tabular}
}
\caption{COMET and Term Recall of every model and language pair. Each cell is the mean over 3 training seeds, along with standard deviation. Support is the number of instances used to score each metric.}
\label{tab:comet-term-recall}
\end{table*}

\subsection{\texorpdfstring{\ref{rq:1}}{RQ1}: Does \sysname\ raise term recall without hurting translation quality}
\label{sec:analysis:rq1}

Table~\ref{tab:comet-term-recall} shows COMET and term recall for four models on seven language pairs. \todo{for bleu see appendix} For every model and language pair, \sysname\ has a greater term recall than \fsftOrig. Compared with \fsftOrig, \sysname\ improves COMET in almost every setting. The gains also hold with OLMo, even though it is pretrained mostly on English text. For example, for Ta-En in OLMo-3-7B, term recall increases from 59.9 to 70.9 with also a one point increase in COMET. 
 Table~\ref{tab:bleu} shows BLEU, in which \sysname\ improves it in all the settings.

\todo{Fix numbering.}
\subsection{\texorpdfstring{\ref{rq:2}}{RQ2}: Which properties of the context-target pairs promote ICTL}
\label{sec:analysis:rq2}

\begin{table*}[t]
\centering
\small
\setlength{\tabcolsep}{3pt}
\renewcommand{\arraystretch}{1.1}
\setlength{\aboverulesep}{0pt}
\setlength{\belowrulesep}{0pt}
\setlength{\extrarowheight}{1.5pt}
\resizebox{\textwidth}{!}{%
\begin{tabular}{lc>{\columncolor{gray!12}}cc>{\columncolor{gray!12}}cc>{\columncolor{gray!12}}cc>{\columncolor{gray!12}}cc>{\columncolor{gray!12}}c}
\toprule
Training & \multicolumn{2}{c}{Llama-3.2-1B / Ha-En} & \multicolumn{2}{c}{Qwen2.5-1.5B / As-En} & \multicolumn{2}{c}{Qwen2.5-1.5B / Ne-En} & \multicolumn{2}{c}{Qwen2.5-3B / Ta-En} & \multicolumn{2}{c}{OLMo-3-7B / En-Ha} \\
\cmidrule(lr){2-3} \cmidrule(lr){4-5} \cmidrule(lr){6-7} \cmidrule(lr){8-9} \cmidrule(lr){10-11}
 & COMET & TermRecall & COMET & TermRecall & COMET & TermRecall & COMET & TermRecall & COMET & TermRecall \\
\midrule
\fsftOrig & 66.64 {\scriptsize(0.12)} & 49.55 {\scriptsize(1.23)} & 67.92 {\scriptsize(0.14)} & 67.75 {\scriptsize(0.95)} & 79.97 {\scriptsize(0.09)} & 58.55 {\scriptsize(1.62)} & 70.44 {\scriptsize(0.16)} & 54.26 {\scriptsize(1.58)} & 74.78 {\scriptsize(0.19)} & 51.70 {\scriptsize(0.20)} \\
\sysname & 68.97 {\scriptsize(0.07)} & 59.35 {\scriptsize(1.73)} & 68.92 {\scriptsize(0.10)} & 75.39 {\scriptsize(1.41)} & 81.00 {\scriptsize(0.40)} & 65.74 {\scriptsize(2.55)} & 72.20 {\scriptsize(0.07)} & 65.11 {\scriptsize(0.76)} & 76.99 {\scriptsize(0.15)} & 57.32 {\scriptsize(0.30)} \\
\quad\textit{\ourNoCipher} & 68.93 {\scriptsize(0.15)} & 55.11 {\scriptsize(1.02)} & 68.34 {\scriptsize(0.03)} & 71.91 {\scriptsize(0.71)} & 81.09 {\scriptsize(0.14)} & 62.60 {\scriptsize(2.81)} & 71.96 {\scriptsize(0.03)} & 63.39 {\scriptsize(0.57)} & 77.01 {\scriptsize(0.15)} & 56.66 {\scriptsize(0.25)} \\
\quad\textit{\ourNoAnneal} & 68.41 {\scriptsize(0.14)} & 59.29 {\scriptsize(1.15)} & 67.19 {\scriptsize(0.34)} & 78.66 {\scriptsize(0.40)} & 80.06 {\scriptsize(0.31)} & 70.64 {\scriptsize(2.14)} & 66.68 {\scriptsize(0.76)} & 70.35 {\scriptsize(1.65)} & 76.76 {\scriptsize(0.15)} & 57.90 {\scriptsize(0.10)} \\
\quad\textit{\ourNoSelect} & 68.76 {\scriptsize(0.15)} & 53.52 {\scriptsize(1.11)} & 68.63 {\scriptsize(0.09)} & 72.49 {\scriptsize(1.55)} & 81.11 {\scriptsize(0.18)} & 65.47 {\scriptsize(0.79)} & 71.77 {\scriptsize(0.20)} & 63.21 {\scriptsize(0.04)} & 76.20 {\scriptsize(0.24)} & 55.15 {\scriptsize(1.21)} \\
\bottomrule
\end{tabular}
}
\caption{Ablations of \sysname{} (Section~\ref{sec:analysis:rq2}). Removing the cipher (\textit{\ourNoCipher}) or the rarity threshold (\textit{\ourNoSelect}) lowers term recall, while removing annealing (\textit{\ourNoAnneal}) lowers COMET. Each cell is the mean over 3 training seeds, along with standard deviation.}
\label{tab:ablation-comet-term-recall}
\end{table*}
\todo{Ablation table too crammed.  Drop on column.}

\sysname\ builds training data from three components: ciphering rare word shared spans, annealing from ciphered to original data (Section~\ref{sec:anneal}), and a rarity threshold $\beta$ that decides which shared spans are selected (Section~\ref{sec:cipher_replacement}). Table~\ref{tab:ablation-comet-term-recall} shows the effect of removing one component at a time.

Ciphering strengthens the model's copying mechanism. \ourNoCipher\ selects the same shared spans as \sysname\ but leaves them unciphered. Term recall drops on every pair. \ourNoCipher\ recalls more rare terms than \fsftOrig{}, so selecting in-context examples that share rare terms with the target helps on its own, and ciphering boosts it. 

Annealing helps to maintain translation quality. \ourNoAnneal\ trains on ciphered data for the whole run. While it has the highest term recall, but its COMET drops on every pair, and falls below \sysname. Training only on ciphered targets teaches the model to copy, but the ciphers distort the target-language text which hampers its ability to produce good quality translations.

\ourNoSelect\ selects to pair based on any span shared by $\vyt$ and an in-context target without checking that they are rare. For frequent words, parametric memory could memorize the term translation.  Ciphering frequent shared words could hurt for two reasons: The original translation task sees more interference, and when ciphering is tapered off via annealing, the original targets may be learned in parametric memory.
We show ablations by varying $\beta$ in Appendix~\ref{sec:appendix:beta_vary}. 

Overall, ciphering and the rarity threshold are what build the ICTL mechanism, and annealing helps keep \sysname\ a good translator.

\subsection{\texorpdfstring{\ref{rq:3}}{RQ3}: Does \sysname\ rely on learning shortcuts}
\label{sec:analysis:rq3}
\label{sec:analysis:robustness}

\begin{figure*}[t]
    \centering
    \includegraphics[width=\textwidth]{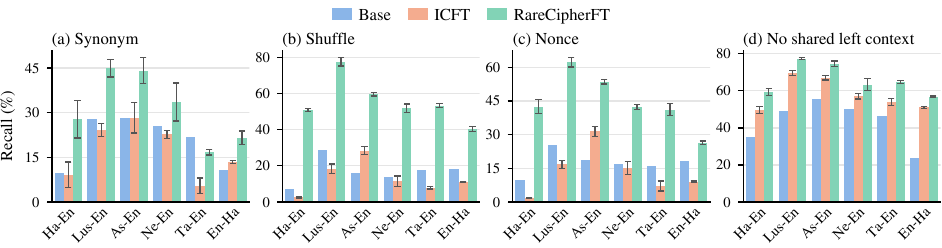}%
    \caption{\label{fig:probe} Term recall under the probes of Section~\ref{sec:analysis:robustness}: (a)~parametric memory (synonym), (b,~c)~semantic similarity (shuffle, nonce), and (d)~without shared left-side context. The models shown are Llama-3.2-1B (Ha-En), OLMo-3-7B (Lus-En, En-Ha), Qwen2.5-1.5B (As-En, Ne-En) and Qwen2.5-3B (Ta-En). Mean over 3 training seeds, error bars are the standard deviation.}
\end{figure*}


Higher term recall alone does not show that a model has learned to take terms from its context via proper ICTL. A model could produce the right term through various shortcuts, and we go through and probe for each in turn.

\xhdr{Recalling from Parametric Memory}
\label{sec:analysis:synonym}
To check that the model has indeed copied a term's translation from the context, instead of recalling from its parameters, we replace every occurrence of the target term in $\vy^i$ with a synonym generated by Gemini 3.8 Flash~\cite{google2026gemini38flash} (prompt in Appendix~\ref{sec:appendix:synonym_prompt}). A model relying on memory keeps producing the original term, while a model that reads the context will produce the synonym.  Figure~\ref{fig:probe} shows that \sysname{} produces the synonym significantly more often than \fsftOrig{}.

\xhdr{Recalling based on Semantic Similarity}
\label{sec:analysis:shuffle_nonce}
If a model aligns target words to their occurrence in the source purely based on their semantic embeddings being similar, the alignment may fail for rare words whose meaning may not yet be well-represented.  For such cases, an alignment that relies on the structural grammar of the two languages could be more useful. 
To isolate this confound, we replace the rare term with a random shuffle of its characters. Shuffled words look like the ciphers that \sysname{} sees during training (Section~\ref{sec:cipher_replacement}), so we also replace rare terms with nonce words, which have English-like pronunciations but no meaning, built from common subwords (e.g.\ \textit{imporear}, \textit{rabbingnest}, \textit{griefini}
). These do not look like anything in the training data. \todo{do fertility analysis}
\sysname{} has higher recall on shuffled words and nonce words (Figure~\ref{fig:probe}) compared to \fsftOrig{}.

\xhdr{Copying from context based on shared prefix}
\label{sec:analysis:context}
Another shortcut a model could take at a decoding step $t$, is to  copy based purely on similarity of the decoded prefix $y^*_1,\ldots,y^*_{t-1}$ somewhere in $\vy^i$, ignoring their source sentences.
To rule out this shortcut, we compute term recall only over rare terms whose preceding words in $\vyt$ do not appear before the correct shared term in any in-context target $\vy^i$. On these terms, \sysname{} still clearly outperforms \fsftOrig{} (Figure~\ref{fig:probe}), so it does not find the term by continuing an n-gram from the context.

%
Together, these results answer \ref{rq:3}: \sysname{} does not rely on these shortcuts, and instead gains by linking the rare source term to its translation in the in-context examples.



\subsection{\texorpdfstring{\ref{rq:5}}{RQ5}: Can raw parallel sentences replace an explicit term dictionary}
\label{sec:analysis:rq5}

\begin{table*}[t]
\centering
\small
\setlength{\tabcolsep}{3pt}
\renewcommand{\arraystretch}{1.1}
\setlength{\aboverulesep}{0pt}
\setlength{\belowrulesep}{0pt}
\setlength{\extrarowheight}{1.5pt}
\resizebox{\textwidth}{!}{%
\begin{tabular}{lc>{\columncolor{gray!12}}cc>{\columncolor{gray!12}}cc>{\columncolor{gray!12}}cc>{\columncolor{gray!12}}cc>{\columncolor{gray!12}}cc>{\columncolor{gray!12}}c}
\toprule
Training & \multicolumn{2}{c}{Llama-3.2-1B / Ha-En} & \multicolumn{2}{c}{OLMo-3-7B / Lus-En} & \multicolumn{2}{c}{Qwen2.5-1.5B / As-En} & \multicolumn{2}{c}{Qwen2.5-1.5B / Ne-En} & \multicolumn{2}{c}{Qwen2.5-3B / Ta-En} & \multicolumn{2}{c}{Llama-3.2-1B / En-Ha} \\
\cmidrule(lr){2-3} \cmidrule(lr){4-5} \cmidrule(lr){6-7} \cmidrule(lr){8-9} \cmidrule(lr){10-11} \cmidrule(lr){12-13}
 & COMET & TermRecall & COMET & TermRecall & COMET & TermRecall & COMET & TermRecall & COMET & TermRecall & COMET & TermRecall \\
\midrule
\sysname & 68.97 {\scriptsize(0.07)} & 69.28 {\scriptsize(1.13)} & 60.23 {\scriptsize(0.38)} & 78.17 {\scriptsize(0.87)} & 68.92 {\scriptsize(0.10)} & 76.72 {\scriptsize(1.35)} & 81.00 {\scriptsize(0.40)} & 67.65 {\scriptsize(0.31)} & 72.20 {\scriptsize(0.07)} & 68.16 {\scriptsize(1.16)} & 75.99 {\scriptsize(0.20)} & 60.22 {\scriptsize(0.44)} \\
DragFT & 68.34 {\scriptsize(0.00)} & 63.40 {\scriptsize(0.75)} & 58.00 {\scriptsize(0.29)} & 82.88 {\scriptsize(0.81)} & 66.66 {\scriptsize(0.04)} & 79.32 {\scriptsize(0.71)} & 80.51 {\scriptsize(0.17)} & 69.50 {\scriptsize(1.53)} & 70.06 {\scriptsize(0.15)} & 67.42 {\scriptsize(0.89)} & 75.29 {\scriptsize(0.08)} & 59.25 {\scriptsize(0.16)} \\
\midrule
Support & 1997 & 102 & 2000 & 184 & 2000 & 483 & 500 & 169 & 1998 & 322 & 1997 & 1384 \\
\bottomrule
\end{tabular}
}
\caption{COMET and Term Recall of \sysname{} and DragFT (Section~\ref{sec:analysis:rq5}). Each cell is the mean over 3 training seeds, along with standard deviation. Support is the number of test instances in each column.}
\label{tab:dragft}
\end{table*}

Table~\ref{tab:dragft} compares \sysname\ with a state-of-the-art dictionary-based MT method.  
DragFT uses a term dictionary to replace source terms with their translations, while \sysname\ uses no dictionary and has to find translations in the ICL examples.  For the low resource languages we were working with, no separate dictionary was available.  So we used DragFT's training dictionary as described in Section~\ref{sec:methods} after filtering on rare words (Appendix~\ref{sec:app:tr}). We evaluate only on instances for which the dictionary edits the test source $\vxt$.

We observe that \sysname\ has higher term recall in half of the settings. More interestingly \sysname\ has higher COMET than DragFT in all settings. This is because DragFT's blind replacement of target words in the source, could fail under source words with ambiguous translations or wrong dictionary entries (Figure~\ref{box:dragft-wrong}). In contrast, \sysname\ learns term translations from the ICL examples where the usage provides greater context, and the model decides whether to copy based on the overall sentence.

\begin{figure}[t]
\begin{promptbox}
\small
\textbf{Source:} \netext{\colorbox{green}{कुसुम} देवी मृतक महेश प्रशादकी प्रथम स्वास्नी थिइन्।}\\
\textbf{DragFT source:} \colorbox{red!20}{shivasgirl} devi deceased mahesh \netext{प्रशादकी प्रथम स्वास्नी थिइन्।}

\vspace{0.5em}

\textbf{Gold:} \colorbox{yellow}{Kusum} Devi was the first wife of deceased Mahesh Prasad.

\textbf{Output (DragFT):} \colorbox{red!20}{shivasgirl} devi was the first wife of mahesh prashad.

\textbf{Output (\sysname):} \colorbox{yellow}{Kusum} Devi was the first wife of Mahesh Prasad.
\end{promptbox}
\caption{\label{box:dragft-wrong}A wrong dictionary entry copied by DragFT, Nepali$\to$English. DragFT's dictionary maps the name \netext{कुसुम} (\emph{Kusum}) to \emph{shivasgirl}.  Transliterations in Figure~\ref{fig:translit-ne}.}
\end{figure}

\section{Conclusion}
\vspace{-0.07in}
We introduced in-context term learning (ICTL)\todo{term or terminology}, the capability of a translation model to learn rare term translations from parallel in-context examples at deployment, without a glossary, an aligner, or any other terminology pipeline. ICTL requires an implicit chain of reasoning: match a rare source term to its occurrence in an in-context source, align it to its translation in the corresponding target, and copy that translation to the right output position. We showed that \fsftOrig\ develops this chain only weakly.

We proposed \sysname, which strengthens ICTL through two changes to the training data, at no extra training cost. First, it pairs each target with in-context examples that share a rare term. Second, it replaces the shared term with a character-shuffled cipher. Both changes lead to settings that are easier to learn by contextual copying than parametric recall. Across seven translation directions and four models from 1B to 7B parameters, \sysname{} raised term recall over \fsftOrig\ by 5–15 points without lowering COMET. Ablations showed that rare-term selection and ciphering each contribute to the gain, while annealing protects translation quality. Probes with synonyms, shuffled and nonce words, and controls for shared surface context indicated that the gains come from linking source terms to their in-context translations rather than from shortcuts. 
We show that methods like DragFT designed to harness an explicit term dictionary produce less fluent output than \sysname.
These results suggest that how training data is composed shapes whether models learn from context, which matters most for rare languages where context may be the only source of new terms.

\section*{Limitations}

\todo{check}

\paragraph{A strong alignment dictionary makes \sysname{} unnecessary.}
\sysname{} teaches the model to infer term translations from in-context examples because we assume no bilingual dictionary covers the test vocabulary. If a high-coverage source--target glossary is available, the translations of rare terms can be given to the model directly. In that case the model no longer needs to recover term alignments from the examples.  In such cases \sysname{} could only be helpful if the entries in the glossary have ambiguous translations. Our method is meant for low-resource language pairs where no such dictionary exists.

\paragraph{The evidence for the mechanism is behavioral.}
Our probes rule out specific shortcuts: parametric recall, semantic similarity and continuing an n-gram from the context, but they do not identify the circuit the model uses, and other shortcuts may remain. The synonym probe uses LLM-generated synonyms, which can occasionally shift meaning or part of speech.

\bibliography{custom,ICL,pubs,custom_cipher}

\appendix

\section{More details of Experiment Setup}
\subsection{Datasets}
\label{app:datasets}
\begin{table}[h]
\centering
\small
\setlength{\tabcolsep}{3pt}
\begin{tabular}{|c|c|ccc|}
\hline
Split & LP & Source & Size & \shortstack{Few-shot\\pool} \\
\hline
\multirow{5}{*}{Train}
 & Ta-En & \href{https://huggingface.co/datasets/ai4bharat/samanantar}{Samanantar} & 15k & 100k \\
 & As-En/En-As & \href{https://data.statmt.org/wmt23/indic-mt/}{IndicNE} & 11k/9k & 25k \\
 & Lus-En & \href{https://data.statmt.org/wmt23/indic-mt/}{IndicNE} & 9k & 25k \\
 & Ne-En & \href{https://aikosh.indiaai.gov.in/home/datasets/details/ishaan_parallel_corpus_1.html}{ISHAAN} & 15k & 100k \\
 & Ha-En/En-Ha & \href{https://huggingface.co/datasets/allenai/wmt22_african}{WMT22-Afr} & 13k/11k & 100k \\
\hline
\multirow{5}{*}{Test}
 & Ta-En & \href{https://ufal.mff.cuni.cz/~ramasamy/parallel/html/}{EnTam} & 2k & 1k \\
 & As-En/En-As & \href{https://data.statmt.org/wmt23/indic-mt/}{IndicNE} & 2k & 2k \\
 & Lus-En & \href{https://data.statmt.org/wmt23/indic-mt/}{IndicNE} & 2k & 1.5k \\
 & Ne-En & \href{https://aikosh.indiaai.gov.in/home/datasets/details/ishaan_parallel_corpus_1.html}{ISHAAN} & 500 & 4.1k \\
 & Ha-En/En-Ha & \href{https://www.statmt.org/wmt21/translation-task.html}{WMT21} & 2k & 2k \\
\hline
\end{tabular}
\caption{Datasets Used for MT. IndicNE = IndicNE-Corp1.0; WMT22-Afr = WMT22 African.
\textbf{Sources:} Samanantar from \citet{samanantar}; EnTam from \citet{biblio:RaBoMorphologicalProcessing2012}; IndicNE-Corp1.0 from \citet{pakray-etal-2024-findings};
ISHAAN from \citet{ishaan-corpus};
WMT22 African from \citet{adelani-etal-2022-findings};
WMT21 from \citet{akhbardeh-etal-2021-findings}.
\label{tab:dataset-mt}
}
\end{table}


\subsection{Fine-tuning details}
\label{sec:app:ft}
For training, we adopted LoRA with a rank of $16$, a scaling factor of $\alpha = 32$, and a dropout rate of $0.05$. Training is performed for a total of 2 epochs (for Ta-En, Ne-En), or 4 epochs (for As-En, Lus-En, Ha-En, En-As, En-Ha
), with a batch size of 2 using Adam with a learning rate of $2 \times 10^{-4}$ and a linear decay learning rate scheduler.
LoRA adapters are trainable on the modules
\begin{enumerate*}
    \item Attention: \texttt{q\_proj, k\_proj, v\_proj, o\_proj}
    \item MLP: \texttt{gate\_proj, up\_proj, down\_proj}
\end{enumerate*}.

The models were trained on NVIDIA B200 GPUs. Training one method on all seven language pairs takes 3.8, 4.4, 5.7 and 9.2 GPU-hours per seed for Llama-3.2-1B, Qwen-2.5-1.5B, Qwen-2.5-3B and OLMo-3-7B, respectively.

\subsection{Term Recall Details}
\label{sec:app:tr}
For each evaluation instance $i$, we construct a candidate set $C_i$ of target-language words, which represent the lexical items that the model is expected to recover from the ICL context rather than from parametric knowledge.
We ensure that $C_i$ satisfies these conditions:
\begin{enumerate*}[leftmargin=*]
    \item The word appears both in the gold reference translation and in the target of at least one ICL example associated with the same prompt.
    \item The word occurs at most 100 times in the training references. For X$\to$En, it has Zipf frequency $<4$, since English words rare in our data can be common in pretraining. En$\to$X skips this, as X is largely unseen in pretraining.
    \item Source exclusion as mentioned in the training data creation in Section~\ref{sec:cipher_replacement}. 
\end{enumerate*}



\section{Varying the Rarity Threshold}
\label{sec:appendix:beta_vary}
\begin{figure*}[t]
    \centering
    \includegraphics[width=\textwidth]{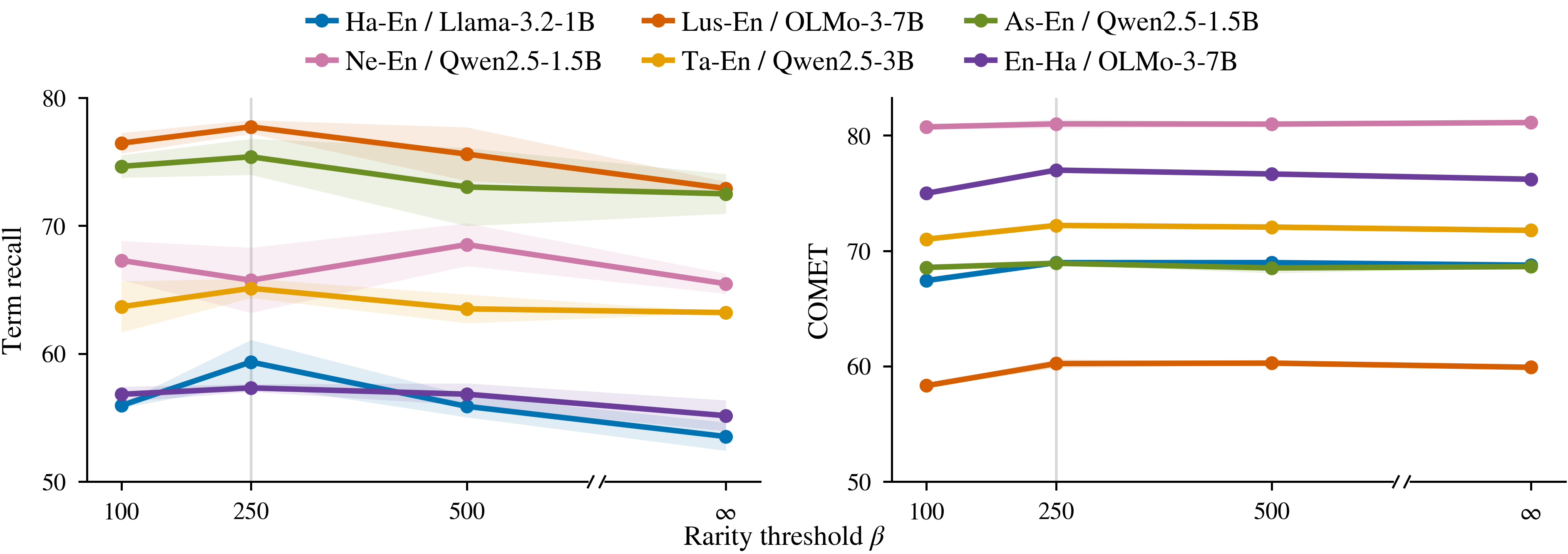}
    \caption{\label{fig:beta_vary} Term recall (left) and COMET (right) as the rarity threshold $\beta$ varies over $\{100, 250, 500, \infty\}$. $\beta=\infty$ removes the threshold (\ourNoSelect). Without the threshold, term recall falls below its $\beta=250$ value for every pair. Mean is taken over 3 training seeds, shading is the standard deviation.
}
\end{figure*}

Figure~\ref{fig:beta_vary} varies the threshold over $\beta \in \{100, 250, 500, \infty\}$. The value used in the main experiments, $\beta=250$, gives the best term recall on almost every pair and COMET barely changes across the sweep. A stricter threshold ($\beta=100$) lowers term recall and gives the lowest COMET on most pairs. This may be due to fewer training instances having a shared rare word span as the threshold is stricter. A moderate threshold works best, and we do not tune it further.

\section{Synonym Generation Prompt}
\label{sec:appendix:synonym_prompt}
The synonyms used in the parametric-memory probe (Section~\ref{sec:analysis:synonym}) are generated once per rare word by Gemini 3.8 Flash~\cite{google2026gemini38flash}, with thinking level set to low and the default temperature, using the prompt in Figure~\ref{fig:synonym_prompt}. The answer is constrained to a JSON object with a single string field. In the prompt, \texttt{\{lang\}} is the target language of the pair (English for X$\to$En, X for En$\to$X), \texttt{\{word\}} is the rare term, and \texttt{\{examples\}} lists up to three in-context target sentences in which it occurs, one per line, each prefixed by ``\texttt{- }'' and followed by its source sentence in square brackets. A proposal is discarded if it is \texttt{NONE}, if it is within the fuzzy-match edit distance of the original word, if it is not itself a rare word under the criterion of Appendix~\ref{sec:app:tr}, or if it already occurs in the instance or among the candidate terms.

\begin{figure*}[t]
\begin{promptbox}
\textbf{System:}
\begin{wrapverb}[fontsize=\footnotesize]
You are a {lang} lexicographer helping build a controlled substitution list for a machine-translation experiment. You replace a {lang} word with a single-word {lang} synonym that a competent translator could have written instead, in the same sentence, with no other edit.
\end{wrapverb}
\textbf{User:}
\begin{wrapverb}[fontsize=\footnotesize]
{lang} word: {word}

Example {lang} sentences where it occurs (source sentence in brackets):
{examples}

Give ONE {lang} word that could replace "{word}" in those sentences without changing the meaning.

Rules:
- Exactly one {lang} word, written in standard {lang} orthography. No spaces, no hyphens, no
  explanation.
- It must be a real {lang} word that fluent speakers use. Never join two words together to make
  one. If you cannot name a real single word, answer NONE.
- It must itself be an UNCOMMON word -- one a fluent reader knows but does not write every day.
  Do not answer with a function word or an everyday high-frequency word.
- Same grammatical form as "{word}" (same number, gender, tense or aspect, part of speech, and
  any attached affix).
- Same register as the examples (legal, medical, religious, journalistic -- match it).
- It must be a genuinely different word, not a spelling variant of "{word}" (not the same word
  with different diacritics or a different spelling).
- Answer NONE if "{word}" is a proper noun, a place, a person, an acronym, a brand, a drug name
  or a number, or if no single-word synonym exists.

Answer with the word alone, or NONE.
\end{wrapverb}
\end{promptbox}
\caption{\label{fig:synonym_prompt} The prompt used to generate a synonym for a rare target term. \texttt{\{lang\}} is the target language of the pair.}
\end{figure*}

\section{BLEU Scores}
\label{sec:appendix:bleu}
Table~\ref{tab:bleu} gives the BLEU score of every model and language pair, for the same runs as Table~\ref{tab:comet-term-recall}.

\begin{table*}[t]
\centering
\small
\setlength{\tabcolsep}{3pt}
\renewcommand{\arraystretch}{1.0}
\setlength{\aboverulesep}{0pt}
\setlength{\belowrulesep}{0pt}
\setlength{\extrarowheight}{0pt}
\begin{tabular}{llccccc}
\toprule
\rule{0pt}{2.3ex}Pair & Training & \multicolumn{1}{c}{Llama-3.2-1B} & \multicolumn{1}{c}{Qwen2.5-1.5B} & \multicolumn{1}{c}{Qwen2.5-3B} & \multicolumn{1}{c}{OLMo-3-7B} & \multicolumn{1}{c}{Support} \\
\midrule
\multirow{3}{*}{Lus-En} & \rule{0pt}{2.3ex}Base & 4.78 & 4.79 & 5.81 & 5.79 & \multirow{3}{*}{2000} \\
 & \fsftOrig & 11.04 {\scriptsize(0.36)} & 12.31 {\scriptsize(0.32)} & 14.15 {\scriptsize(0.20)} & 15.64 {\scriptsize(0.32)} &  \\
 & \sysname & \cellcolor{green!25}\textbf{11.53} {\scriptsize(0.29)} & \cellcolor{green!25}\textbf{12.54} {\scriptsize(0.76)} & \cellcolor{green!25}\textbf{14.43} {\scriptsize(0.41)} & \cellcolor{green!25}\textbf{16.36} {\scriptsize(0.21)} &  \\
\midrule
\multirow{3}{*}{As-En} & \rule{0pt}{2.3ex}Base & 8.95 & 8.15 & 11.18 & 11.04 & \multirow{3}{*}{2000} \\
 & \fsftOrig & 11.30 {\scriptsize(0.16)} & 10.71 {\scriptsize(0.14)} & 13.56 {\scriptsize(0.14)} & 14.67 {\scriptsize(0.21)} &  \\
 & \sysname & \cellcolor{green!25}\textbf{11.68} {\scriptsize(0.24)} & \cellcolor{green!25}\textbf{11.44} {\scriptsize(0.27)} & \cellcolor{green!25}\textbf{14.30} {\scriptsize(0.30)} & \cellcolor{green!25}\textbf{15.27} {\scriptsize(0.18)} &  \\
\midrule
\multirow{3}{*}{Ta-En} & \rule{0pt}{2.3ex}Base & 5.27 & 3.95 & 6.48 & 9.32 & \multirow{3}{*}{1998} \\
 & \fsftOrig & 7.67 {\scriptsize(0.10)} & 7.34 {\scriptsize(0.18)} & 9.26 {\scriptsize(0.11)} & 11.77 {\scriptsize(0.07)} &  \\
 & \sysname & \cellcolor{green!25}\textbf{8.82} {\scriptsize(0.07)} & \cellcolor{green!25}\textbf{8.59} {\scriptsize(0.03)} & \cellcolor{green!25}\textbf{10.84} {\scriptsize(0.14)} & \cellcolor{green!25}\textbf{13.28} {\scriptsize(0.14)} &  \\
\midrule
\multirow{3}{*}{Ne-En} & \rule{0pt}{2.3ex}Base & 14.77 & 9.98 & 15.42 & 19.95 & \multirow{3}{*}{500} \\
 & \fsftOrig & 17.85 {\scriptsize(0.33)} & 16.69 {\scriptsize(0.23)} & 20.29 {\scriptsize(0.38)} & 21.67 {\scriptsize(0.26)} &  \\
 & \sysname & \cellcolor{green!25}\textbf{19.25} {\scriptsize(0.12)} & \cellcolor{green!25}\textbf{18.54} {\scriptsize(0.09)} & \cellcolor{green!25}\textbf{21.80} {\scriptsize(0.02)} & \cellcolor{green!25}\textbf{23.32} {\scriptsize(0.24)} &  \\
\midrule
\multirow{3}{*}{Ha-En} & \rule{0pt}{2.3ex}Base & 5.73 & 4.33 & 5.03 & 6.56 & \multirow{3}{*}{1997} \\
 & \fsftOrig & 10.14 {\scriptsize(0.08)} & 10.06 {\scriptsize(0.18)} & 10.89 {\scriptsize(0.19)} & 12.54 {\scriptsize(0.26)} &  \\
 & \sysname & \cellcolor{green!25}\textbf{11.72} {\scriptsize(0.09)} & \cellcolor{green!25}\textbf{11.82} {\scriptsize(0.24)} & \cellcolor{green!25}\textbf{12.77} {\scriptsize(0.29)} & \cellcolor{green!25}\textbf{13.97} {\scriptsize(0.16)} &  \\
\midrule
\multirow{3}{*}{En-As} & \rule{0pt}{2.3ex}Base & 4.20 & 3.62 & 4.44 & 4.99 & \multirow{3}{*}{2000} \\
 & \fsftOrig & 8.40 {\scriptsize(0.12)} & 7.72 {\scriptsize(0.08)} & 8.53 {\scriptsize(0.01)} & 9.99 {\scriptsize(0.07)} &  \\
 & \sysname & \cellcolor{green!25}\textbf{8.68} {\scriptsize(0.06)} & \cellcolor{green!25}\textbf{8.04} {\scriptsize(0.24)} & \cellcolor{green!25}\textbf{9.07} {\scriptsize(0.11)} & \cellcolor{green!25}\textbf{10.11} {\scriptsize(0.12)} &  \\
\midrule
\multirow{3}{*}{En-Ha} & \rule{0pt}{2.3ex}Base & 2.77 & 3.04 & 2.93 & 2.99 & \multirow{3}{*}{1997} \\
 & \fsftOrig & 10.74 {\scriptsize(0.10)} & 10.64 {\scriptsize(0.10)} & 11.13 {\scriptsize(0.04)} & 11.45 {\scriptsize(0.15)} &  \\
 & \sysname & \cellcolor{green!25}\textbf{11.81} {\scriptsize(0.08)} & \cellcolor{green!25}\textbf{11.93} {\scriptsize(0.12)} & \cellcolor{green!25}\textbf{12.31} {\scriptsize(0.12)} & \cellcolor{green!25}\textbf{12.58} {\scriptsize(0.22)} &  \\
\bottomrule
\end{tabular}
\caption{BLEU of every model and language pair. Each cell is the mean over 3 training seeds, along with standard deviation. Support is the number of instances used to score each metric.}
\label{tab:bleu}
\end{table*}

\section{Anecdotes}
\label{sec:appendix:anecdotes}
\begin{figure*}[t]
\begin{promptbox}

Consider the following 5 translations from Mizo to English.

\textbf{Example 1}\\
\textbf{Source:} Hunte a liam tam poh leh \colorbox{green}{Jehova ngaihdam} theih ph\^{a}k b\^{a}k niin ka inngai tawlh tawlh a.\\
\textbf{Target:} ``The more time that went by, the more I thought I was beyond \colorbox{yellow}{Jehovah's forgiveness}.''

\vspace{0.5em}

\textbf{Example 2}\\
\textbf{Source:} An fate chu t\^{u}n hma ang kh\^{a}n an awm leh ang a , Ka hmaah ng\^{e}lnghet takin an awm ang . Anmahni hnehchhiahtute chu ka hrem z\^{e}l ang .\\
\textbf{Target:} Their children also shall be as before , and their congregation shall be established before me ; and I will punish all who oppress them .

\vspace{0.5em}

\textbf{Example 3}\\
\textbf{Source:} Ka t\^{a}n puithiam rinawm tak ka din thar leh ang a , chu chuan ka duhz\^{a}wng takin thil a ti ang . An chh\^{u}ngkua nghet takin ka din ang a , ka hriakthiha hmaah chuan rawng a b\^{a}wl reng ang ,\\
\textbf{Target:} I will raise me up a faithful priest , that shall do according to that which is in my heart and in my mind . I will build him a sure house ; and he shall walk before my anointed forever .

\vspace{0.5em}

\textbf{Example 4}\\
\textbf{Source:} I ngilneihna a hlun d\^{a}wn tih ka hria , I rinawmzia lah v\^{a}n anga nghet a ni .\\
\textbf{Target:} I indeed declare , `` Love stands firm forever . You established the heavens . Your faithfulness is in them . ''

\vspace{0.5em}

\textbf{Example 5}\\
\textbf{Source:} 11:1 ) Hm\^{a}n laia Jehova mite chu an rinna tinghet reng t\^{u}rin eng thil dangin nge \d{t}anpui?\\
\textbf{Target:} 11:1) Prayer also played a vital part in keeping the faith of pre-Christian witnesses strong.

\vspace{0.5em}

Translate the source text from Mizo to English.

\textbf{Source:} An sim a , \colorbox{green}{Jehova ngaihdamna}ah rinna nghet tak an nei bawk .

\vspace{0.5em}

\textbf{Gold:} They are repentant , and they have abiding faith in \colorbox{yellow}{Jehovah's forgiveness} .

\textbf{Output (\fsftOrig):} They have faith in Jehovah and in his purpose.

\textbf{Output (\sysname):} They have a firm faith in \colorbox{yellow}{Jehovah's forgiveness}.
\end{promptbox}
\caption{\label{box:anecdote2}Terminology copying from context, Mizo$\to$English. The source term is highlighted in \colorbox{green}{green} and its English rendering in \colorbox{yellow}{yellow}. The stem \emph{ngaihdam} in the test source also occurs in Example~1, whose target supplies \emph{forgiveness}; \sysname\ copies it and otherwise differs from the reference only in dropping the opening clause, while \fsftOrig\ replaces the term with the vaguer \emph{in his purpose}.}
\end{figure*}

An example in Figure~\ref{box:anecdote2} shows how our model does a better job at using terminology-specific words from the context in Mizo, where the baseline instead paraphrases the term away. It can be observed that the English word `forgiveness', the translation of the Mizo stem `ngaihdam', is copied from the in-context examples, and our model's generation differs from the reference only in dropping the opening clause. In contrast, \fsftOrig\ replaces the term with the vaguer `in his purpose', which is fluent and topically apt but carries none of the lexical content of the source.

\section{Transliterations of Non-Latin Examples}
\label{app:translit}
\par\noindent\begin{minipage}{\columnwidth}
\begin{promptbox}
\small
\textbf{$\vx^1$:} \textit{tēciya a\ub{r}iviyal ti\ub{n}attaiyoṭṭi, viññā\ub{n}ikaḷukku \ldots}\\
\textbf{$\vx^2$:} \textit{\ldots{} ve\ub{r}\ub{r}i karamāka ēviya \ldots}\\
\textbf{$\vxt$:} \textit{inta cāta\ub{n}aiyai pārāṭṭi piratamar viññā\ub{n}ikaḷukku tolaipēci mūlamāka vā\ub{l}ttu terivittuḷḷār.}
\end{promptbox}
\captionof{figure}{\label{fig:translit-ta}Transliteration of the Tamil sources in Figure~\ref{fig:example}.}
\end{minipage}
\vspace{1em}

\par\noindent\begin{minipage}{\columnwidth}
\begin{promptbox}
\small
\textbf{Source:} \textit{kusum devī mṛtak maheś praśādkī pratham svāsnī thiin.}\\
\textbf{DragFT source:} \textit{shivasgirl devi deceased mahesh praśādkī pratham svāsnī thiin.}
\end{promptbox}
\captionof{figure}{\label{fig:translit-ne}Transliteration of the Nepali sources in Figure~\ref{box:dragft-wrong}.}
\end{minipage}

\end{document}